\documentclass[conference]{IEEEtran}
\IEEEoverridecommandlockouts

\usepackage{times}
\usepackage{amssymb}
\usepackage{amsmath}
\usepackage{amsthm}
\usepackage{graphics}
\usepackage{epsfig}
\usepackage{multirow}
\usepackage{nicefrac}
\usepackage{cases}          
\usepackage{subcaption}
\usepackage{mathtools}
\usepackage{xcolor}
\usepackage{algorithm}
\usepackage{algorithmicx}
\usepackage{algpseudocode}
\usepackage{balance}
\usepackage{gensymb}
\usepackage{comment}
\usepackage{hyperref}
\usepackage[capitalise]{cleveref}
\usepackage{bm}

\DeclareMathAlphabet{\pazocal}{OMS}{zplm}{m}{n}

\newcommand{\mat}[0]{\begin{bmatrix}}
\newcommand{\mate}[0]{\end{bmatrix}}
\newcommand{\va}{\mathbf{a}}

\newcommand{\vs}{\mathbf{s}}
\newcommand{\vS}{\mathbf{S}}
\newcommand{\vo}{\mathbf{o}}

\newcommand{\vu}{\mathbf{u}}

\newcommand{\vx}{\mathbf{x}}

\newcommand{\cA}{\mathcal{A}}

\newcommand{\cS}{\mathcal{S}}

\newcommand{\cU}{\mathcal{U}}

\newcommand{\cX}{\mathcal{X}}

\newcommand\norm[1]{\left\|#1\right\|}              

\newcommand{\la}{\leftarrow}

\newtheorem{definition}{\textbf{\textit{Definition}}}

\begin{document}
%
\title{Learning Interaction-aware Guidance Policies for Motion Planning in Dense Traffic Scenarios}
%
%
%


\author{Bruno~Brito$^1$,
        Achin~Agarwal$^1$,
        and~Javier~Alonso-Mora$^1$
\thanks{This work was supported by the Amsterdam Institute for Advanced Metropolitan Solutions and the Netherlands Organisation for Scientific Research (NWO) domain Applied Sciences (Veni 15916).}
\thanks{$^1$The authors are with the Cognitive Robotics (CoR) department,
        Delft University of Technology, 2628 CD Delft, The Netherlands
    {\tt\small \{bruno.debrito, j.alonsomora\}@tudelft.nl}}
}

%
%

\markboth{IEEE TRANSACTIONS ON INTELLIGENT TRANSPORTATION SYSTEMS, VOL. XXX, NO. XXX, XXX 2021}%
{Shell \MakeLowercase{\textit{et al.}}: Bare Demo of IEEEtran.cls for IEEE Journals}
%





\maketitle

\begin{abstract}
Autonomous navigation in dense traffic scenarios remains challenging for autonomous vehicles (AVs) because the intentions of other drivers are not directly observable and AVs have to deal with a wide range of driving behaviors. 
To maneuver through dense traffic, AVs must be able to reason how their actions affect others (interaction model) and exploit this reasoning to navigate through dense traffic safely.
This paper presents a novel framework for interaction-aware motion planning in dense traffic scenarios. We explore the connection between human driving behavior and their velocity changes when interacting. Hence, we propose to learn, via deep Reinforcement Learning (RL), an interaction-aware policy providing global guidance about the cooperativeness of other vehicles to an optimization-based planner ensuring safety and kinematic feasibility through constraint satisfaction. The learned policy can reason and guide the local optimization-based planner with interactive behavior to pro-actively merge in dense traffic while remaining safe in case the other vehicles do not yield. 
We present qualitative and quantitative results in highly interactive simulation environments (highway merging and unprotected left turns) against two baseline approaches, a learning-based and an optimization-based method. The presented results demonstrate that our method significantly reduces the number of collisions and increases the success rate with respect to both learning-based and optimization-based baselines.

%
\end{abstract}

\begin{IEEEkeywords}
Deep reinforcement learning, dense traffic, motion planning, safe learning, trajectory optimization.
\end{IEEEkeywords}

%
\IEEEpeerreviewmaketitle

\section{Introduction}

\begin{figure}[t]
\centering
\includegraphics[scale=0.26]{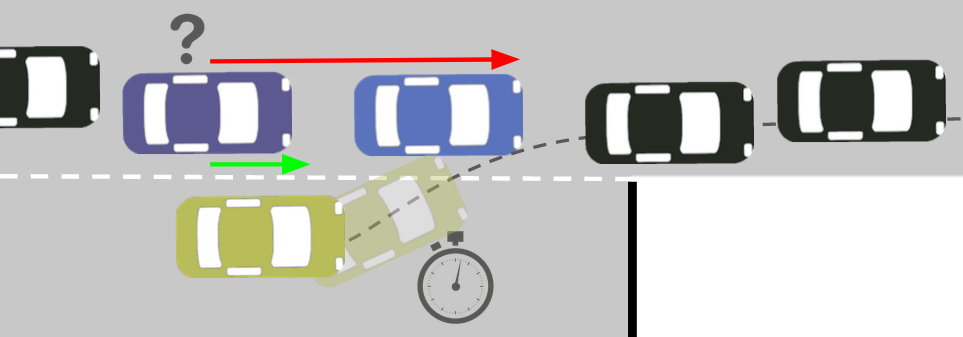}

\caption{Illustration of a dense on-ramp merging traffic scenario where the autonomous vehicle (yellow) needs to interact with other traffic participants in order to merge onto the main lane in a timely and safe manner. The potential follower (purple) may yield (green arrow) to the autonomous vehicle leaving space for the autonomous vehicle to merge or behave non-cooperatively (red arrow) to deter the autonomous vehicle from merging. To successfully merge, the autonomous vehicle needs to identify the cooperative ones by interacting with them without any explicit inter-vehicle communication.
}
\label{fig:first}
\end{figure}

\IEEEPARstart{D}{espite} recent advancements in autonomous driving solutions (e.g., Waymo \cite{Bansal2019}, Uber \cite{Sadat2020}), driving in real-world dense traffic scenarios such as highway merging and unprotected left turns still stands as a hurdle in the widespread deployment of autonomous vehicles (AVs) \cite{Schwarting2018}. Driving in dense traffic conditions is intrinsically an interactive task \cite{ulbrich2015structuring}, where the AVs' actions elicit immediate reactions from nearby traffic participants and vice-versa. An example of such behavior is illustrated in Fig. \ref{fig:first}, where the autonomous vehicle needs to perform a merging maneuver onto the main lane.
To accomplish this task, it needs to first reason about the other driver's intentions (e.g., to yield or not to yield) without any explicit inter-vehicle communication. Then, it needs to know how to interact with multiple road-users and leverage other vehicles' cooperativeness to induce them to yield, such that they create room for the AV to merge safely.

\IEEEpubidadjcol

The development of interaction-aware prediction models has been studied \cite{rudenko2020human,mozaffari2020deep}, allowing AVs to reason about other drivers' intentions. In contrast, developing interactive motion planning algorithms that can reason and exploit other drivers cooperativeness is still challenging \cite{Schwarting2019}. The majority of traditional motion planning methods are too conservative and fail in dense scenarios because they do not account for the interaction between the autonomous vehicle and nearby traffic \cite{Schwarting2018}, \cite{Gonzalez2016}. However, works that account for the interaction among agents do not scale for many agents due to the curse of dimensionality \cite{Schmerling2018,cleac2020lucidgames,Trautman2018}. Deep Reinforcement Learning (DRL) methods can overcome the latter, but either do not provide any safety guarantees \cite{Saxena2019} or are overly conservative to ensure safety \cite{bouton2019reinforcement}.

In this paper, we introduce an interactive Model Predictive Controller (IntMPC) for safe navigation in dense traffic scenarios. We explore the insight that human drivers communicate their intentions and negotiate their driving maneuvers by adjusting both distance and time headway to the other vehicles \cite{zgonnikov2020should,toledo2003modeling}. Studies show that in dense traffic scenarios, such as merging and left-turning, 
cooperative or aggressive behavior is strongly connected to higher or smaller average distance and time headway \cite{marczak2013key,davis2004field}, respectively. 
These driving features (i.e., relative distance and time headway) can be directly translated into a velocity reference. Hence, we propose to learn, via Deep Reinforcement Learning (DRL), an interaction-aware policy as a velocity reference. This reference provides global guidance to a local optimization-based planner, which ensures that the generated trajectories are kino-dynamically feasible and safety constraints are respected.
Our method leverages vehicles' interaction effects to create free-space areas for the AV to navigate and complete various driving maneuvers in cluttered environments.  

    
    

\section{Related Work}

The literature devoted to the problem of modeling human interactions among traffic participants is vast \cite{Schwarting2018} and includes 
rule-based, optimization-based, game theoretic and learning-based methods.

\subsection{Rule-based Methods}
Traditional autonomous navigation systems typically employ a sequential planning architecture hierarchically decomposing the planning and decision-making pipeline into different blocks such as perception, behavioral planning, motion planning and low-level control \cite{paden2016survey}. Rule-based methods translate implicit and explicit human-driving rules into handcrafted functions directly influencing the motion planning phase. These methods have demonstrated excellent ability to solve specific problems (e.g., precedence at an intersection followed by waiting for the availability of enough free space for the vehicle to pass safely) \cite{Baker2008, Urmson2008, Montemerlo2008}. Nevertheless, these methods do not consider the interactions between multiple traffic participants and thus can fail in dense traffic scenarios.

\subsection{Search-based Methods}
The decision-making problem for autonomous navigation is inherently a Partially Observable Markov Decision Process (POMDP) because the other drivers' intentions are not directly observable but can be estimated from sensor data \cite{bai2015intention}. To improve decision-making and intention estimation, it has been proposed to incorporate the road context information \cite{7225835}. To deal with a variable number of agents, dimensional reduction techniques have been employed to create a compressed and fixed-size representation of the other agents information \cite{Hubmann2017}. Yet, solving a POMDP online can become infeasible if the right assumptions on the state, action and observation space are not made. For instance, \cite{Hubmann2018} proposed to use Monte Carlo Tree Search (MCTS) algorithms to obtain an approximate optimal solution online and \cite{Zhou2018} improved the interaction modeling by proposing to feedback the vehicle commands into planning. These methods demonstrated promising results but are limited to environments for which they were specifically designed, demand high computational power and can only consider a discrete set of actions. 

\subsection{Optimization-based Methods}
Optimization-based methods are widely used for motion planning since they allow to define collision and kino-dynamics constraints explicitly. These methods include receding-horizon control techniques which allow to plan in real-time and incorporate predicted information by optimizing over a time horizon \cite{Schwarting2018,ferranti2019safevru,park2019planner}. However, these works employ simple prediction models and do not consider interaction. Recently, data-driven methods allow to generate interaction-aware predictions \cite{brito2020social} that can be used for planning \cite{bae2020cooperation,ivanovic2020mats}. However, these methods ignore the influence of the ego vehicle's actions in the planning phase struggling to find a collision free trajectory in highly dense traffic scenarios \cite{Trautman2010}. Not only motion planners must account for the interaction among the driving agents but also generate motions plans which respect social constraints. Hence, to generate socially compatible plans, Inverse Optimal Control techniques have been used to learn human-drivers preferences  \cite{Sadigh2016}, \cite{You2019AdvancedLearning}. These methods either fail to scale to interact with multiple agents \cite{Sadigh2016} or can only handle a discrete set of actions \cite{You2019AdvancedLearning} rendering them incapable to be used safely in highly interactive and dense traffic scenarios.

\subsection{Game Theoretic Methods}

Game Theoretic approaches such as \cite{Fisac2019} model the interaction among agents as a game allowing to infer the influence on each agent's plans. However, the task of modeling interactions requires the inter-dependency of all agents on each other's actions to be embedded within the framework. This results in an exponential growth of interactions as the number of agents increases, rendering the problem computationally intractable. Social value orientation (SVO) is a psychological metric used to classify human driving behavior. \cite{Schwarting2019} models the interaction problem as a dynamic game given the other driver`s SVO. Similarly, a unscented Kalman filter is used to iteratively update an estimate of the other drivers`cost parameters \cite{cleac2020lucidgames}. Nevertheless, these approaches require local approximations to find a solution in a tractable manner. Cognitive hierarchy reasoning \cite{Camerer2004AGames} allows to reduce the complexity of these algorithms by assuming that an agent performs a limited number of iterations of strategic reasoning. For instance, iterative level-k model based on cognitive hierarchy reasoning \cite{Camerer2004AGames} has been used to obtain a near optimal policy for performing merge maneuvers \cite{Garzon2019GameScenarios} and lane change \cite{bouton2020reinforcement} in highly dense traffic scenarios. However, these approaches consider a discrete action space and do not scale well with the number of vehicles. 

\subsection{Learning-based Methods}

Learning-based approaches leverage on large data collection to build interaction-aware prediction models \cite{brito2020social} or to learn a driving policy directly from sensor data \cite{bojarski2016end}. For instance, generative adversarial networks can be used to learn a driving policy imitating human-driving behavior \cite{kuefler2017imitating}. Conditioning these policies on high-level driving information allows to use it for planning \cite{codevilla2018end}. Moreover, to account for human-robot interaction these policies can be conditioned on the interaction history \cite{Schmerling2018}. Yet, the deployment of these models can lead to catastrophic failures when evaluated in new scenarios or if the training dataset is biased and unbalanced \cite{amini2018variational}. 

Reinforcement Learning (RL) has shown great potential for autonomous driving in dense traffic scenarios \cite{Wu2017FlowAA,Bouton2019}. For example, DQN has been employed to learn negotiating behavior for lane change \cite{Wang2018,mukadam2017tactical} and intersection scenarios \cite{tram2018learning}. Yet, the latter consider a discrete and limited action space. In contrast, in \cite{Saxena2019} it is proposed to learn a continuous policy (jerk and steering rate) allowing to achieve smooth control of the vehicle. These methods are able to learn a working policy under highly interactive traffic conditions involving multiple entities. However, they fail to provide safety guarantees and reliability, rendering these methods vulnerable to collisions. Recently, a vast amount of works has proposed different ways to introduce safety guarantees of learned RL policies \cite{hewing2020learning}. The key idea behind these works is to synthesize a safety controller when an unsafe action is detected by employing formal verification methods \cite{fulton2018safe}, computing offline safe reachability sets \cite{fisac2019bridging} or employing safe barrier functions \cite{Cheng2020}. To reduce conservativeness, \cite{bouton2019reinforcement} proposes to use Linear Temporal Logic to enforce safety probabilistic guarantees.
However, \emph{safe RL} methods do not account for interaction among the agents, being highly conservative in dense environments. Finally, close to our work, \cite{8916922} learned a decision-making policy to select from a discrete and limited set of predefined constraints which ones to enable in an MPC formulation and thus, controlling the vehicle behavior applied to intersection scenarios. In contrast, we propose to learn a continuous interaction-aware policy providing global guidance to an MPC through the cost function.


\subsection{Contribution}

The main contributions of this work are:
\begin{itemize}
    \item An Interactive Model Predictive Controller (IntMPC) for navigation in dense traffic environments combining DRL to learn an interaction-aware policy providing global guidance (velocity reference) in the cost function to a local optimization-based planner;
    \item Extensive simulation results demonstrate that our approach triggers interactive negotiating behavior to reason about the other drivers' cooperation and exploit their cooperativeness to induce them to yield while remaining safe.
\end{itemize}


\begin{figure*}[!t]
  \centering
  \begin{minipage}{\textwidth}
    \includegraphics[height=4cm,width=18cm,trim={0cm 0cm 0cm 0cm},clip]{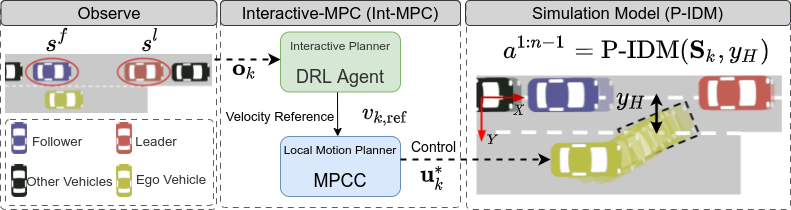}
 
  \end{minipage}

  \caption{
  Our proposed architecture comprises of three main modules: an Interactive Reinforcement Learner (DRL Agent), a Local Motion Planner (MPCC) and a Simulation Model (P-IDM). The AV observes the leader state $s^l$ and follower state $s^f$ relative to it, which serves as input to the Interactive Planner providing a reference velocity $v_{k,\textrm{ref}} = \pi(\vs_k,\vS_k) $ for the MPCC to follow. The MPCC then computes locally optimal sequence of control commands $\mathbf{u}_{0: H-1}^{*}$ minimizing a cost function $J(\mathbf{\vs_k}, \vu_{k})$ (See \cref{sec:mpc}). The reference velocity $v_{\textrm{ref}}$ allows to directly control the AV aggressiveness and thus, to control the interaction with the other vehicles. Finally, P-IDM then computes acceleration command for the other vehicles based on the estimated AV's motion plan (\cref{sec:abl_simulation_models}).}
  \label{fig:overview}
\end{figure*}

\section{Preliminaries}
Throughout this paper, vectors are denoted in bold lowercase letters, $\mathbf{x}$, matrices in capital, $ M $, and sets in calligraphic uppercase, $\mathcal{S}$.  $\norm{\vx} $ denotes the Euclidean norm of $ \vx$ and $\norm{\vx}_Q = \vx^TQ\vx$ denotes the weighted squared norm. Variables $\{\vs,\va\}$ denote the state and action used in the RL formulation, and $\vu$ denotes the control command for the AV.

\subsection{Problem Formulation}

Consider a set $\mathcal{X}$ of $n$ vehicles interacting in a dense traffic scenario comprising an autonomous vehicle (AV) and $n-1$ human drivers, henceforth referred to as other vehicles, exhibiting different levels of willingness to yield. The term "vehicles" is used to collectively refer to the AV and other vehicles. At the beginning of an episode, the AV receives a global reference path $\pazocal{P}$ to follow from a path planner consisting of a sequence of $M$ waypoints $\mathbf{p}^{r}_m = [x^r_m,y^r_m] \in \mathbb{R}^2$ with $m \in \pazocal{M} := \{1,\dots,M\}$.
For each time-step $k$, the AV observes its state $\vs_k$ and the states of other agents $\vS_k=[ \vs^1_k ,\dots,\vs^{n-1}_k]$, then takes action $\va_k$, leading to the immediate reward $R(\vs_k,\va_k)$ and next state $\vs_{k+1}=f(\vs_k,\vu_k)$, under the dynamic model $f$\footnote{This is identical to the Vehicle Model used in the simulation defined in \cref{sec:vehicle_model}} and controller model $h$, with $\vu_k=h(\vs_k,\va_k)$. 
The vehicle's state is defined as

\begin{equation*}
\mathbf{s}^{i}_k=\{x_k, y_k, \psi_k, v_k\}  \forall i \in \{0, \ldots, n-1\}
\end{equation*}
where $x_k$ and $y_k$ are the Cartesian position coordinates, $\psi_k$ the heading angle and $v_k$ the forward velocity in a global inertial frame $\mathcal{W}$ fixed in the main lane (see \cref{fig:overview}). 
 $\pazocal{A}^{\textrm{ego}}$ and $\pazocal{A}^{i}$ denote the area occupied by the AV and the $i$-th other vehicle, respectively. 
We aim to learn a continuous policy $\pi(\va_k|\vs_k,\vS_k)$ conditioned on the AV's and other vehicles' states 
minimizing the expected driving time $\mathbb{E}[t_g]$ for the AV to reach its goal position while ensuring collision-free motions, defined as the following optimization problem:

\begin{subequations}\label{eqn:cost}
\begin{align}
\pi^* = & \underset{\pi}{\operatorname{argmin}} \quad  \mathbb{E}\left[t_{g} \mid  \pi(\va_k|\vs_k,\vS_k)\right] \nonumber\\
\text { s.t. } \quad &\mathbf{s}_{k+1}=f(\mathbf{s}_{k}, \mathbf{u}_{k}), \label{eq:dynamic_constraints}\\
& \vu_k = h(\vs_k,\pi(\va_k|\vs_k,\vS_k)) \\
& \pazocal{A}^{\textrm{ego}}_k\cap \pazocal{A}^{i}_k=\emptyset \label{eq:collision_constraints}\\
&\label{eq:state_control_constraints} \vu_k \in \cU, \,\, \vs_{k} \in \cS, \,\, \va_{t} \in \cA, \\
& \forall i \in \{1\dots n-1\} \,\,\, \forall k \in \{0\dots t_g\} \nonumber
\end{align}
\end{subequations}
where (\ref{eq:dynamic_constraints}) are the kino-dynamic constraints, (\ref{eq:collision_constraints}) the collision avoidance constraints, and $\cS$, $\cA$ and $\cU$ are the set of admissible states, actions, and control inputs (e.g., maximum vehicles' speed), respectively.
We assume that each vehicle's current position and velocity are observed (e.g., from on-board sensor data) and no inter-vehicle communication.


\section{Interactive Model Predictive Control}



\subsection{Overview}
This section introduces the proposed Interactive Model Predictive Control (IntMPC) framework for safe navigation in dense traffic scenarios. Figure \ref{fig:overview} depicts our proposed motion planning architecture incorporating three main modules: an interactive reinforcement learner, a local optimization planner (MPCC), and an interactive simulation environment. Firstly, we define the RL framework 
to learn an interaction-aware navigation policy (Section \ref{sec:rl}), providing global guidance to a local optimization planner (Section \ref{sec:mpc}). Secondly, we introduce the behavior module used to simulate dense traffic scenarios with various driving behavior, ranging from cooperative to non-cooperative. Here, we propose an expansion for the Intelligent Driver Model (IDM) model allowing the other vehicles to react to the other's predicted plans (Section\ref{sec:other_agents_model}). To finalize, we introduce our training algorithm to jointly train the interaction-aware policy and the local optimization planner (Section\ref{sec:training_alg}). Our IntMPC enhances the AV with interactive behavior, exploiting the other traffic participants’ interaction effects.

\subsection{Interactive Planner}\label{sec:rl}
Here, we propose to use deep RL to learn an interaction-aware velocity reference exploiting the interaction effects between the vehicles and providing global guidance to a local optimization-based planner.

\subsubsection{RL Formulation}
The AV's observation vector is composed by the leader's (vehicle in front) and the follower's (vehicle behind the AV) state, $\mathbf{o}_k=[\mathbf{s}^{l}_k,\mathbf{s}^{f}_k]$, relative to AV's frame.
To enable interactive behavior with the other traffic participants, we define the RL policy's action as a velocity reference to directly control the interaction at the merging point. High-speed values lead to more aggressive and low-speed to more conservative behavior, respectively.
Hence, we consider a continuous action space $\mathcal{A}\subset \mathbb{R}$ and aim to learn the optimal policy $\pi$ mapping the AV's state and observation to a probability distribution of actions.

\begin{subequations}
\begin{align}
    \pi_\theta(\mathbf{s}_k,\mathbf{o}_k) = \mathbf{a}_k = v_{k,\textrm{ref}} \label{eq:policy}\\
    \pi_\theta(\mathbf{s}_k,\mathbf{o}_k) \sim  \mathcal{N}(\mu_k,\sigma_k)
\end{align}
\end{subequations}
where $\theta$ are the policy's network parameters, $\mathcal{N}$ is a multivariate Gaussian density function, and $\mu$ and $\sigma$ are the Gaussian's mean and variance, respectively. 



We formulate a reward function to motivate progress along a reference path, to penalize collisions and infeasible solutions, and when moving too close to another vehicle. The reward function is the summation of the four terms described as follows:

\begin{equation}
R\left(\mathbf{s}_k,\mathbf{o}_k, \mathbf{a}_k\right)=\left\{\begin{array}{ll}
v_k  &  \\

r_{\textrm{infeasible}} & \text { if } c_{k}^{c, i} > 1 \,\, \\

r_{\textrm{collision}} & \text { if } \pazocal{A}^{\mathrm{ego}}_k\cap \pazocal{A}^{\mathrm{i}}_k\neq \emptyset \\

r_{\textrm{near}} & d_{\textrm{min}}(\mathbf{s}_k,\mathbf{s}^i_k) \le \Delta d_{\textrm{min}} \,\, \\
\end{array}\right.
\end{equation}
where $c_{k}^{c, i}$ is the collision avoidance constraint between the AV and the vehicle $i$ (\cref{sec:dynamic_collision})
, $\pazocal{A}^{ego}_k\cap \pazocal{A}^{i}_k$ represents the common area occupied by the AV and the $i$-th other vehicle at step $k$. $d_{\textrm{min}}$ is the minimum distance to the closest nearby vehicle $i$ and $\Delta d_{\textrm{min}}$ is a hyper-parameter distance threshold. 
The first term $v_k$ is a reward proportional to the AV's velocity encouraging higher velocities and thus, encouraging interaction and minimizing the time to goal. The second $r_{\textrm{infeasible}}$, third $r_{\textrm{collision}}$ and fourth term $r_{\textrm{near}}$ penalize the AV for infeasible solutions, collisions and for driving too close to other vehicles, respectively.


\subsection{Local Motion Planner}\label{sec:mpc}
Deep RL can be used to learn an end-to-end control policy in dense traffic scenarios \cite{Saxena2019}, \cite{Bouton2019}. However, their sample inefficiency \cite{yu2018towards} and transferability issues \cite{zhao2020sim} makes it hard to apply them in real-world settings. In contrast, optimization-based methods have been widely used and deployed into actual autonomous vehicles \cite{Schwarting2018a,ferranti2019safevru}. Therefore, we employ Model Predictive Contour Control (MPCC) to generate locally optimal control commands following a reference path while satisfying kino-dynamics and collision avoidance constraints if a feasible solution is found. The reference path can be provided by a global path planner such as Rapidly-exploring Random Trees (RRT) \cite{cl-rrt}.

\subsubsection{Vehicle Model}
\label{sec:vehicle_model}
We employ a kinematic bicycle model for the AV, described as follows:

\begin{equation} \label{eqn:vehicle_model}
    \begin{array}{l}
\dot{x}=v \cos (\phi+\beta) \\
\dot{y}=v \sin (\phi+\beta) \\
\dot{\phi}=\dfrac{v}{l_{r}} \sin (\beta) \\
\dot{v}=u^{a} \\
\beta=\arctan \left(\dfrac{l_{r}}{l_{f}+l_{r}} \tan \left(u^{\delta}\right)\right)
\end{array}
\end{equation}
where $\beta$ is the velocity angle. The distances of the rear and front tires from the center of gravity of the vehicle are $l_{r}$ and $l_{f}$, respectively, and are assumed to be identical for simplicity. The vehicle control input $\mathbf{u}$ is the forward acceleration $u^{a}$ and steering angle $u^{\delta}$, $\mathbf{u}=[u^{a},u^{\delta}]$.

\subsubsection{Cost Function}
The local controller receives a velocity reference $v_\textrm{ref}$, from the Interactive Planner (\cref{sec:rl}), exploiting for the interaction effects of the AV in the other vehicles to maximize long-term rewards. To enable the AV to follow the reference path while tracking the velocity reference, we define the stage cost as follows:


\begin{equation} \label{eqn:lossfunction}
    \begin{aligned}
J(\mathbf{\mathbf{\vs}_k}, \mathbf{u}_{k},\lambda_{k} ) &=\norm{e_{k}^{c}(\mathbf{\vs}_k,\lambda_{k})}_{q_c}+\norm{e_{k}^{l}(\mathbf{\vs}_k,\lambda_{k})}_{q_l} \\
&+\norm{v_{k,ref} - v_{k}}_{q_v}+\norm{u_{k}^{a}}_{q_u}+\norm{u_{k}^{\delta}}_{q_\delta}
\end{aligned}
\end{equation}
where $\mathcal{Q}=\{q_{c}, q_{l},q_{v},q_{u}, q_{\delta}\}$ denotes the set of cost weights and $\lambda_{k}$ is the estimated progress
along the reference path. 
To track the reference path closely, we minimize two cost terms: the contour error ($e_{k}^{c}$) and lag error ($e_{k}^{l}$). Contour error gives a measure of how far the ego vehicle deviates from the reference path laterally whereas lag error measures the deviation of the ego vehicle from the reference path in the longitudinal direction.
For more details on path parameterization and tracking error, please refer to \cite{ferranti2019safevru}. The third term, $\|v_{k,\textrm{ref}} - v_{k}\|$, motivates the planner to follow $v_{\textrm{ref}}$ closely. Finally, to generate smooth trajectories, we add a quadratic penalty to the control commands $u_{k}^{a}$ and $u_{k}^{\delta}$.

\subsubsection{Dynamic Obstacle Avoidance} \label{sec:dynamic_collision}
The occupied area by the AV, 
$\cA^{\mathrm{ego}}(\boldsymbol{s}_k)$, is approximated with a union of $n_c$ circles i.e $\bar{A}^{\mathrm{ego}}(\boldsymbol{s}_k) \subseteq \bigcup_{c \in\left\{1, \ldots, n_{c}\right\}} \mathcal{A}_{c}(\boldsymbol{s}_k)$, where $\mathcal{A}_{c}$ is the area occupied for a circle with radius $r$. For each vehicle $i$, the occupied area $\mathcal{A}^{i}$ is approximated by an ellipse of semi-major axis $a_i$, semi-minor axis $b_i$ and orientation $\phi$. 
To ensure collision-free motions, we define a set of non-linear constraints imposing that each circle $c$ of the AV with the elliptical area occupied by the $i$-th vehicle does not intersect:
\begin{equation}\label{eq:ineq_constraint}
\begin{split}
\!\!\!c_k^{i,c}(\mathbf{s}_k,\mathbf{s}^i_k) \!\!= \!\!\begin{bmatrix}
\Delta x_k^c\\
\Delta y_k^c
\end{bmatrix}^\textrm{T}\!\!\!\!
R(\phi)^\textrm{T}\!\!\begin{bmatrix}
\frac{1}{\alpha^2} & 0\\ 0 & \frac{1}{\beta^2}
\end{bmatrix}\!\! R(\phi)\!\!\!\ \begin{bmatrix}
\Delta x_k^c\\
\Delta y_k^c
\end{bmatrix}
> 1,
\end{split}
\end{equation}
$\forall k \in\{0, \ldots, H\}$ and $\forall i \in\{1, \ldots, n-1\}$. The parameters $\Delta x_{k}^{c}$ and $\Delta y_{k}^{c}$ represent x-y relative distances 
in AV's frame between the disc $c$ and the ellipse $i$ for prediction step $k$. To guarantee collision avoidance we enlarge the other vehicle's semi-major and semi-minor axis with a $r_{\textrm{disc}}$ coefficient, assuming $\alpha = a + r_{\textrm{disc}}$ and $\beta = b + r_\textrm{disc}$ as described in \cite{Brito2019}.

\subsubsection{Road boundaries}
We introduce constraints on the lateral distance (i.e., contour error) of the AV with respect to the reference path to ensure that the vehicle stays within the road boundaries \cite{Zhou2018}:
\begin{equation}
    -w_{\textrm{left}}^{\textrm{road}} \leq e_{k}^{c}(\vs_k) \leq w_{\textrm{right}}^{\textrm{road}}
    \label{eqn:road_boundary}
\end{equation}
where $w_{\textrm{left}}^{\textrm{road}}$ and $w_{\textrm{right}}^{\textrm{road}}$ are the left and right load limits, respectively.

\subsubsection{MPC Formulation}
We formulate the motion planning problem as 
a Receding Horizon Trajectory Optimization problem (\ref{eq:mpc_formulation}) with planning horizon $H$ conditioned on the following constraints:
\begin{subequations}\label{eq:mpc_formulation}
\begin{align}
\mathbf{u}_{0: H-1}^{*}= &\min _{u_{0: H-1}} \sum_{k=0}^{H-1} J(\vs_k, \mathbf{u}_{k},\lambda_{k} )+J(\vs_H,\lambda_{H}) \label{eq:mpc_cost} \\
\text { s.t. } \quad & \vs_{k+1}=f(\vs_{k}, \mathbf{u}_{k}), \label{eq:model_constraints} \\
& \lambda_{k+1} = \lambda_{k} + v_k\Delta t \\
& -w_{\textrm{left}}^{\textrm{road}} \leq e^c(\mathbf{s}_k) \leq w_{\textrm{right}}^{\textrm{road}} \label{eq:road_constraints} \\
& c_k^{i,c}(\mathbf{s}_k,\mathbf{s}^i_k) > 1 \hspace{0.5em} \forall c \in \{1,\dots,n_c\}, \label{eq:col_constraints} \\
  & \mathbf{u}_{k} \in \mathcal{U}, \quad
                            \mathbf{s}_k \in \mathcal{S}, \label{eq:state_constraints} \\
    &                       \forall k\in \{0,\dots,H\}.
\end{align}
\end{subequations}
where $\Delta t$ is the discretization time and $\mathbf{u}_{0: H-1}^{*}$ the locally optimal control sequence for H time-steps. 
 In this work, we assume a constant velocity model to estimate of the other vehicles' future positions, as in \cite{Brito2019}.

\subsection{Modeling Other Traffic Drivers' Behaviors}\label{sec:other_agents_model}

We aim to simulate dense and complex negotiating behavior with varying degrees of willingness to yield. For instance, in a typical dense traffic scenario (e.g., on-ramp merging), human drivers trying to merge onto the main lane need to leverage other drivers' cooperativeness to create obstacle-free space to merge safely. In contrast, drivers on the main lane exhibit different levels of willingness to yield. Some drivers stop as soon as they spot the other vehicle on the adjacent lane (Cooperative). Other drivers ignore the other vehicles entirely and may even accelerate to deter it from merging (Non-Cooperative).
Moreover, they also consider an internal belief about the other vehicle's motion plan on the adjacent lane in their decision-making process at the merging point. Here, we introduce the Predictive 
Intelligent Driver Model (P-IDM) to control the longitudinal driving behavior of the other vehicles, built on the Intelligent Driver Model (IDM) \cite{Treiber2000}. Our proposed model consists of three main steps: leader and follower selection, other vehicles' motion estimation, and control command computation. 

\subsubsection{Leader \& Follower Selection}
\begin{figure}[!t]
\centering
\includegraphics[scale=0.27]{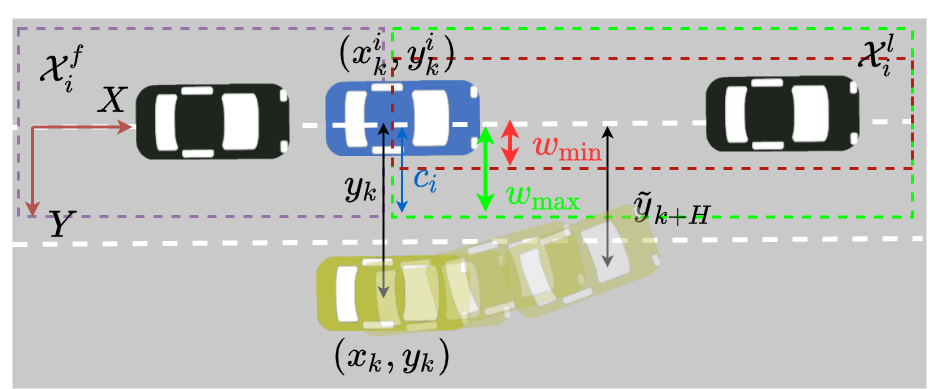}

\caption{ Leader \& Follower Selection Process. The AV is depicted in yellow, the $i$-th interacting vehicle in blue, and the $i$-th vehicle's follower and leader in black. $(x_k,y_k)$ are the x-y position coordinates in the main lane frame of the AV and and $(x^i_k,y^i_k)$ of the $i$-th vehicle on the main lane at time-step $k$. Dashed purple represents the followers' area set. Dashed red and green represent the leader's area set. To model mixed driving behavior, the $i$-th vehicle cooperation coefficient $c_i$ is randomly sampled from a uniform bounded distribution $c_i\sim\mathcal{U}([w_{\mathrm{min}}, w_{\mathrm{max}}])$ (defined in \cref{sec:scenarios}). $w_{\mathrm{max}}$ and $w_{\mathrm{min}}$ represents a maximum and minimum distance between the center of the current lane and the adjacent lane.}
\label{fig:leader_selection}
\end{figure}

For each vehicle, the model assigns a leader, denoted with up-script $l$, and a follower, denoted with up-script $f$. Consider $\mathcal{X}^l_i$ as the set of potential leaders for the vehicle $i$, then 


\begin{definition}[]\label{def:leaders_set}
    IDM: A set $\mathcal{X}^l_i \subseteq \cX$ is the set of possible leaders for the vehicle $i$ if $\forall j \in [0,n-1], j \neq i : x^j_k > x^i_k$ and $y^i_k < c_i$.
\end{definition}
\begin{definition}[]\label{def:followers_set}
    IDM: A set $\mathcal{X}^f_i \subseteq \cX$ is the set of possible followers for the vehicle $i$ if $\forall j \in [0,n-1], j \neq i : x^j_k < x^i_k$ and $y^i_k < c_i$.
\end{definition}
where 
$c_i$ is a hyper-parameter threshold used to model cooperation (\cref{sec:scenarios}) \cite{Saxena2019}. \cref{fig:leader_selection} shows an example of the leader's and follower's sets for the merging scenario as well as the physical representation of the cooperation coefficient $c_i$. In the IDM, the leaders' and followers' sets are defined based on the vehicle's current lateral position, $y^i_k$, leading to reactive behavior. In contrast, we propose to define the leader's and follower's sets based on the estimated lateral position at time-step $H$, $\tilde{y}^i_H$, as it follows
\begin{definition}[]\label{def:leaders_set_new}
    P-IDM: A set $\mathcal{X}^l_i \subseteq \cX$ is the set of possible leaders for the vehicle $i$ if $\forall j \in [0,n-1], j \neq i : x^j_k < x^i_k$ and $\norm{\tilde{y}^i_H} < c_i$.
\end{definition}
Employing the predicted lateral position $\tilde{y}_H$ instead of the current lateral position $y_k$ allows to elicit non-reactive behavior from the other vehicles. The leader for the vehicle $i$ is defined as it follows

\begin{definition}[]\label{def:leader_agent}
    A vehicle $j\in\mathcal{X}^l_i$ is the leader of vehicle $i$ if $ \forall m \in \mathcal{X}^l_i, m \neq j: \norm{x^j_k-x^i_k} \leq \norm{x^m_k-x^i_k}$.
\end{definition}

To model mixed driving behavior, 
$c_i$ is sampled from a uniform bounded distribution $c_i\sim\mathcal{U}([w_{\mathrm{min}}, w_{\mathrm{max}}])$ (defined in \cref{sec:scenarios}). $w_{\mathrm{max}}$ and $w_{\mathrm{min}}$ represents a maximum and minimum distance between the center of the current lane and the adjacent lane, as depicted in \cref{fig:leader_selection}, respectively.
Moreover, the $c_i$ values' range plays an essential role in the final policy's behavior by controlling the proportion of cooperative and non-cooperative vehicles encountered by the AV during training resulting in a more aggressive or conservative final policy.

\subsubsection{Motion Plan Estimation}

To enhance the IDM model with predictive driving behavior, we propose to condition the IDM on the beliefs of the other drivers' motion plans. Specifically, we assume that each vehicle on the main lane maintains an internal belief about the AV's motion plan (on the adjacent lane)\footnote{For the Ramp Merging scenario (detailed in Sec. \ref{merging_scenario}), the current lane corresponds to the main lane whereas the adjacent lane refers to the merge lane whereas for the Unprotected Left Turn scenario (detailed in Section \ref{unprotected_left_scenario}), the current lane refers to the top lane and the adjacent lane corresponds to the bottom lane.}. To estimate the AV's motion plans, different prediction models can be employed (e.g., constant velocity model). Later, in \cref{sec:abl_simulation_models}, we investigate our method's performance for different prediction models.


\subsubsection{Control Command Computation}\label{sec:idm_acc}
For each time-step $k$ and for each vehicle $i$, the acceleration control is computed depending on the vehicle's velocity $v^i_k$ and current distance to the leader $\Delta x^i_k = \norm{(x^i_k,y^i_k)-(x^l_k,y^l_k)}$:

\begin{equation}\label{eq:idm_acc}
u^{a,i}_{k}=a_{\textrm{max}}\left[1-\left(\frac{v^i_{k}}{v^*}\right)^{4}-\left(\frac{s^{*}\left(v^i_{k}, \Delta v^i_{k}\right)}{\Delta x^i_k}\right)^{2}\right]
\end{equation}
where $s^*$ is the desired minimum gap, $a_{\textrm{max}}$ the maximum acceleration, $\Delta v^i_{k}=v^i_k-v^l_k$ the $i$-th vehicle approach rate to the preceding vehicle, and $v^*$ the desired velocity. Please note that we only do longitudinal control for the other vehicles on the main lane by employing \cref{eq:idm_acc}. For the AV, we employ a local optimization-based planner (\cref{sec:mpc}) for steering and acceleration control. 




\subsection{Training Procedure}\label{sec:training_alg}

\begin{algorithm}[t]
\caption{Training Procedure}\label{alg:training}
\begin{algorithmic}[1]
 \State \textbf{Inputs:} planning horizon $H$, initial policy's parameters $\theta$, Q-functions' parameters $\{\phi_1,\phi_2\}$, number of training episodes $n_{\textrm{episodes}}$, number of vehicles $n$,  reward function $R(\mathbf{s}_k,\mathbf{o}_k,\va_{k})$ and number of control steps $K$
 
 \State Initialize initial states: $\{\mathbf{s}_0,\dots,\mathbf{s}^{n-1}_0\}\sim \cS $
 \State Initialize replay buffer: $\mathcal{D} \la \emptyset$
 \While{$episode < n_{\textrm{episodes}}$}
 \State Get observation $\mathbf{o}_k$ and AV's state $\mathbf{s}_k$
 
 \If{$k\bmod K == 0$}
 \State Sample velocity reference for the AV: 
 
 $v_{k,\textrm{ref}}\sim \pi_{\theta}(\mathbf{s}_k,\mathbf{o}_k)$

 \EndIf
 \State Solve the optimization problem of Eq.\ref{eq:mpc_formulation} without collision constraints (Eq.\ref{eq:col_constraints})  : 
 
 $\mathbf{u}_{k:k+H}^{*}$ = MPCC$(v_{k,\textrm{ref}}, \mathbf{s}_{k},\mathbf{o}_{k})$
  
  \State Estimate AV's lateral position: 
  
  $\tilde{y}_{H} = $ PredictionModel$(v_{k}, \mathbf{s}_{k},\mathbf{o}_{k})$ (\cref{sec:abl_simulation_models})
  
  \For{$i \in \{1,\dots, n-1\} $}
  
  \State Select leader for vehicle $i$ (\cref{def:leaders_set_new} and \cref{def:leader_agent})
  
  \State Compute longitudinal acceleration for the $i$-th vehicle:
  
  $u^{a,i}_k = $P-IDM($\vs^i_k,\vs^l_k$) (Section \ref{sec:idm_acc}): 
  
  \EndFor
  
  \State Observe next vehicles' states $\{\mathbf{s},\dots,\mathbf{s}^{n-1}\}$, reward $r_k$ and done signal
  
  \State Store $(\mathbf{s}_k,\va_k,r_k,\mathbf{s}_{k+1},\textrm{done})$ in replay buffer $ \mathcal{D}$
  
  \If{done}
            \State $episode\ +=1$
             \State Initialize: $\{\vs_0,\dots,\vs^n_0\}\sim \cS$ 
        \EndIf
 \If{it's time to update}
    \State SAC training \cite{Haarnoja2018}
 \EndIf
 \EndWhile
\State \Return $\{\theta,\phi_1,\phi_2\}$
\end{algorithmic}
\end{algorithm}

In this work, we train the policy using Soft Actor-Critic (SAC) \cite{Haarnoja2018} to learn the policy's probability distribution parameters. SAC augments traditional RL algorithms' objective with the policy's entropy, embedding the notion of exploration into the policy while giving up on clearly unpromising paths \cite{Haarnoja2018}. 
We propose to jointly train the guidance policy with the local motion planner allowing the trained policy to directly implement our method on a real system and learn with the cases resulting in infeasible solutions for the optimization solver. In contrast to prior works on safe RL \cite{Cheng2020}, during training, we do not employ collision constraints (Eq.\ref{eq:col_constraints}), exposing the policy to dangerous situations or collisions which is necessary to learn how to interact with other vehicles closely.


\cref{alg:training} describes the proposed training strategy. 
Each episode begins with the initialization of all vehicle's states (see \cref{sec:scenarios,sec:driving_scenarios} for more details). Every $K$ cycles, we sample a reference velocity $v_\textrm{ref}$ from the policy $\pi_\theta$. Querying the interaction-aware policy every $K$ control cycles helps to stabilize the training procedure and better assess the policy's impact on the environment (see \cref{sec:ablation_k}). 
Then, the MPCC computes a locally optimal sequence of steering and acceleration commands $u^{*}_{0:H-1}$ for the AV. If a feasible solution is found, we apply the first control command of the sequence and re-compute the motion plan in the next cycle considering new observations. If no feasible solution is found, we apply a braking command. 
Afterward, the P-IDM computes an action for each vehicle on the main lane while being aware of the AV on the adjacent lane. An episode is over if: the AV reaches the goal position (finishes merging or turning left); the AV collides with another vehicle; it does not finish the maneuver in time (i.e., timeout).
Finally, to update the policy's distribution parameters, we employ the Soft Actor-Critic (SAC) \cite{Haarnoja2018} method. We refer the reader to \cite{Haarnoja2018} for more details about the learning method's equations. Please note that our approach is agnostic to which RL algorithm we use.

\subsection{Online Planning}

\begin{algorithm}[t]
\caption{Int-MPC}\label{alg:intmpc}
\begin{algorithmic}[1]
 \State \textbf{Inputs:} AV's state $\vs_k$, observation $\vo_k$ and reference path $\mathbf{p}^{r}_m = [x^r_m,y^r_m] \in \mathbb{R}^2$ with $m \in \pazocal{M} := \{1,\dots,M\}$ waypoints.
 
 
 \For{$k = 0,1,2,...$}
 \State Get observation $\mathbf{o}_k$ and AV's state $\mathbf{s}_k$
 \State Sample velocity reference for the AV: 
 
 $v_{k,\textrm{ref}}=\pi_{\theta}(\mathbf{s}_k,\mathbf{o}_k)$
  
  
  \State Compute MPCC trajectory by solving Eq.\ref{eq:mpc_formulation}: 
 
 $\vu_{k:k+H}^{*}$ = MPCC$(v_{k,\textrm{ref}}, \mathbf{s}_{k},\mathbf{o}_{k})$
  
  \If{$\vu_{k:k+H}^{*}$ is \textit{feasible} }
  \State$\text{Apply } \vu^*_k$
  \Else
  \State$\text{Apply } \vu_\textrm{safe}$
  \EndIf
    \EndFor
\end{algorithmic}
\end{algorithm}

\cref{alg:intmpc} describes our Interactive Model Predictive Controller (IntMPC) algorithm. For every step $k$, we first obtain a velocity reference, $v_{\textrm{ref}}$, from the trained policy. Then, by solving the MPCC problem (\cref{eq:mpc_formulation}), we obtain a locally optimal sequence of control commands $\vu^*_{k:k+H}$. Finally, if the MPCC plan is feasible we employ the first control command, $\vu^*_k$, and re-compute a new plan considering the new observations following a receding horizon control strategy. Else, we apply a braking command, $\vu_\textrm{safe}$.

\section{Experiments}

This section presents simulation results for two dense traffic scenarios (\cref{sec:driving_scenarios}) considering different cooperation settings for the other vehicles (\cref{sec:scenarios}). First, we provide an ablation study analyzing our method's design choices (\cref{sec:ablation}). After, we present qualitative (\cref{sec:qualitative}) and performance results (\cref{sec:quantitative}) of our approach against two baselines:

\begin{itemize}
\item DRL : state-of-the-art Deep Reinforcement Learning
approach, SAC \cite{Haarnoja2018}, learning a continuous policy controlling the AV's forward velocity. 
    
\item MPCC \cite{ferranti2019safevru}: Model Predictive Contour Controller with a constant velocity reference.

\end{itemize}

All controller parameters were manually tuned to get the best possible performance.

\subsection{Experimental setup}\label{sec:experiment_setup}
Simulation results were carried out on an Intel Core i9, 32GB of RAM CPU @ 2.40GHz taking approximately 20 hours to train, approximately 20 million simulation steps. The non-linear and non-convex MPCC problem of \cref{eq:mpc_formulation} was solved using the ForcesPro~\cite{zanelli2020forces} solver. Our simulation environment, P-IDM, builds on an open-source highway simulator \cite{Leurent2018} expanding it to incorporate complex interaction behavior. Hyperparameters values can be found in Table \ref{tab:hyperparameters}. Our motion planner and simulation environment are open source\footnote{\url{https://github.com/tud-amr/highway-env}}.


\begin{table}[t]
\begin{center}
\caption{SAC's Hyperparameters}
\label{tab:hyperparameters}
\begin{tabular}{l l}
\hline
 Hyperparameter & Value \\
 \hline
 Planning Horizon & 1.5 s \\
 Number of Stages $N$ & 15 \\
 Number of parallel workers & 7 \\
 Q neural network model & 2 dense layers of 256 \\
 Policy neural network model & 2 dense layers of 256 \\
 Activation units & Relu \\
 Training batch size & 2048 \\  
 Discount factor & 0.99 \\
 Optimizer & Adam \\
 Initial entropy weight ($\alpha$) & 1.0 \\
 Target update ($\tau$) & $5 \times 10^{-3}$  \\
 Target entropy lower bound & -1.0 \\
 Target network update frequency & 1 \\
 Learning rate &  $3 \times 10^{-4}$  \\
 Replay buffer size & $10^6$ \\
 $r_{\textrm{infeasible}}$ & -1 \\
 $r_{\textrm{collision}}$ & -300 \\
 $r_{\textrm{near}}$ & -1.5 \\
 \hline
 \end{tabular}

 \end{center}
\end{table}

\subsection{Driving scenarios}\label{sec:driving_scenarios}
We consider two densely populated driving scenarios: merging on a highway and unprotected left turn. The vehicles are modeled as rectangles with 5 m length and 2 m width. For each episode, the initial distance between the other vehicles is drawn from a uniform distribution ranging from [7, 10] m. Their initial and target velocities are sampled from a uniform distribution, $v^{0:n}_0\sim\mathcal{U}(3,4) $ m/s. This initial configuration prevents early collisions while ensuring no gaps of more than 2 meters \cite{ni2016vehicle}, typical of dense traffic scenarios. These scenarios compel the AV to leverage other vehicles' cooperativeness while also exposing it to a myriad of critical scenarios for the final policy's performance.

\subsubsection{Ramp Merging} \label{merging_scenario}
\begin{figure}
  \centering
\begin{subfigure}{\linewidth}
  \centering
\includegraphics[width=\textwidth]{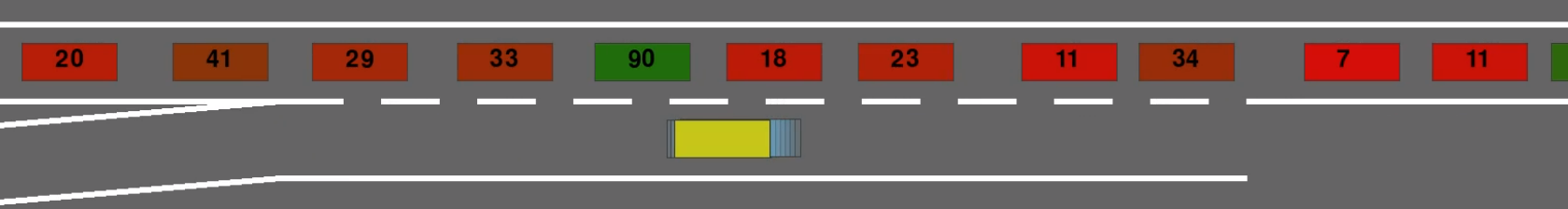}

\caption{Ramp merging scenario. The AV on the main road, bottom lane, has to merge into the main top lane.}
\label{fig:merging}
\end{subfigure}

\begin{subfigure}{\linewidth}
  \centering
\includegraphics[width=\textwidth]{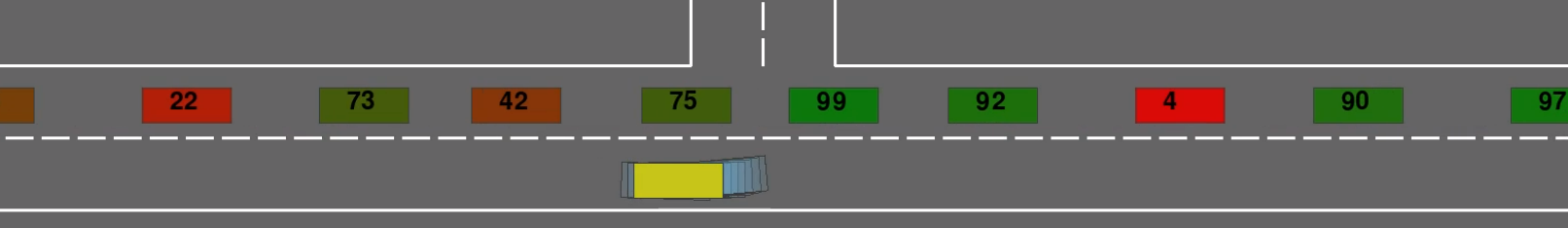}

\caption{Unprotected left-turn scenario: The AV on the main road, bottom lane, has to make a left-turn while avoiding collision with the other vehicles on the main road, top lane.}
\label{fig:left_tu}
\end{subfigure}

  \caption{ Evaluation environments: The AV is depicted in yellow and each other vehicle is assigned with a color transitioning from red (i.e., non-cooperative) to green (i.e., cooperative). The number displayed by each vehicle represents its cooperation coefficient.
  }
  \label{fig:scenarios}
\end{figure}

\cref{fig:merging} depicts an instance of the merging scenario. It comprises two lanes: the main lane and a merging lane. At the beginning of each episode, the main lane is populated with the other vehicles, moving from left to right. In contrast, the merge lane only includes the AV.

\subsubsection{Unprotected Left Turn}
\label{unprotected_left_scenario}
\cref{fig:left_tu} illustrates the unprotected left turn scenario. It consists of two roads: the main road and the left road perpendicular to each other. The main road is populated with the other vehicles (on the top lane) and the AV (on the bottom lane). The other vehicles move from right to left on the main road, whereas the AV is initialized at the bottom lane of the main road, and its objective is to take an unprotected left turn onto the left road.


\subsection{Evaluation Scenarios}\label{sec:scenarios}

We present simulation results considering different settings for the other vehicles' cooperation coefficient:


\begin{itemize}
    \item Cooperative: In this scenario, most vehicles are cooperative ($c^i\sim \mathcal{U}(2,4)$ m), implying that as soon as the AV shows intentions of merging into the main lane, the other vehicle starts considering the AV as its new leader, leaving space for it to merge into the main lane. This evaluation scenario helps in assessing the merging speed of the policy.
    
    \item Non-Cooperative: This scenario comprises mostly non-cooperative vehicles ($c^i\sim \mathcal{U}(0,2)$ m), meaning that the other vehicles would not stop for the AV unless the AV's lateral horizon state is in the top lane. This scenario explicitly assesses the policy's aggressiveness. In these scenarios, the best option for the AV is to stop and wait for gaps and then merge in as quickly as possible.  

    \item Mixed: This traffic scenario involves agents with varying degrees of cooperativeness ($c^i\sim \mathcal{U}(0,4)$ m), featuring a continuous transition from cooperative to non-cooperative vehicles. Here, the goal is to assess how differently the AV behaves with cooperative and non-cooperative vehicles.

\end{itemize}

During training, we consider a mixed setting for the other vehicles.
Rule based methods such as IDM, MOBIL fail in highly dense traffic conditions and thus have not been included for evaluation purposes \cite{Saxena2019}.

\subsection{Evaluation metrics}
\label{evaluation_metrics}

To evaluate our proposed method, we employ the following evaluation metrics:

\begin{itemize}
    \item \textit{Success Rate}: Percentage of successful episodes. An episode is deemed successful if the AV is able to merge on to the main highway or perform a left term without colliding and before timeout.
    
    \item \textit{Collisions}: Percentage of episodes resulting in collision.
    
    \item \textit{Timeout}: Percentage of episodes in which the AV did not reach the goal within the maximum specified time. This metric does not include those episodes that resulted in collision.
    
    \item \textit{Time-to-goal}: Time in seconds for the AV to reach the goal position.
    

\end{itemize}

\subsection{Performance analysis}\label{sec:ablation}

This section investigates the impact of two critical design choices for our proposed approach: MPCC's parameters 
and using a different number of control cycles per RL policy query. Moreover, we evaluate our method's robustness to different prediction models used by the other vehicles to estimate the AV's motion plans.

\subsubsection{Local controller parameters}

\begin{table}[!t]
\begin{center}
\caption{Ablation study of the MPCC's parameters considering a mixed setting for the other vehicles.}
\label{tab:mpcc_weigth}
\begin{tabular}{l c c c}

\hline
 \multicolumn{4}{c}{Success (\%) / Collision (\%) / Timeout (\%)}  \\
 \hline
 & $q_v=0.1$ & $q_v=1.0$ & $q_v=10.0$  \\
 \hline

 $v_{\textrm{ref}}=$ 2 m/s & 58 / 26 / 16   & \textbf{62 / 22 / 16} & 58 / 34 / 8  \\
 $v_{\textrm{ref}}=$ 3 m/s & 48 / 26 / 26 &  62 / 29 / 9  &  55 / 42 / 3    \\
 $v_{\textrm{ref}}=$ 4 m/s & 56 / 25 / 19 & 57 / 35 / 8  & 53 / 46 / 1 \\
 \hline
 \end{tabular}
 \end{center}
\end{table}

\begin{table*}[!t]
  \caption{Sensitivity analysis of the hyperparameter $K$, i.e., number of control cycles per RL policy query, on the learned policy's performance. All policies were trained considering a mixed setting of other vehicles. Querying the RL policy for a new velocity for each two control cycles leads to the best performance (bold values).}
         \centering
\begin{tabular}{c|c|c|c|c|c|c|c|c|c}
\hline
\multicolumn{1}{|c||}{}                 & \multicolumn{3}{|c||}{Cooperative} & \multicolumn{3}{c||}{Mixed} &
\multicolumn{3}{c|}{Non Cooperative} \\  \hline

\multicolumn{1}{|c||}{} &
\multicolumn{1}{c|}{Success(\%)} & \multicolumn{1}{c|}{Collision(\%)} &
\multicolumn{1}{c||}{Timeout(\%)} &
\multicolumn{1}{c|}{Success(\%)} &
\multicolumn{1}{c|}{Collision(\%)} &
\multicolumn{1}{c||}{Timeout(\%)} &
\multicolumn{1}{c|}{Success(\%)} &
\multicolumn{1}{c|}{Collision(\%)} &
\multicolumn{1}{c|}{Timeout(\%)} \\ \hline \hline 

\multicolumn{1}{|c||}{K = 1} &
\multicolumn{1}{c|}{80.0} & 
\multicolumn{1}{c|}{0.0} &
\multicolumn{1}{c||}{20.0} &
\multicolumn{1}{c|}{70.0} &
\multicolumn{1}{c|}{0.0} &
\multicolumn{1}{c||}{30.0} &
\multicolumn{1}{c|}{33.0} &
\multicolumn{1}{c|}{0.0} &
\multicolumn{1}{c|}{67.0} \\ \hline

\multicolumn{1}{|c||}{K = 2} &
\multicolumn{1}{c|}{\textbf{88.0}} & 
\multicolumn{1}{c|}{0.0} &
\multicolumn{1}{c||}{\textbf{12.0}} &
\multicolumn{1}{c|}{\textbf{72.0}} & 
\multicolumn{1}{c|}{0.0} &
\multicolumn{1}{c||}{\textbf{28.0}} &
\multicolumn{1}{c|}{\textbf{37.0}} &
\multicolumn{1}{c|}{0.0} &
\multicolumn{1}{c|}{\textbf{63.0}} \\ \hline

\multicolumn{1}{|c||}{K = 3} &
\multicolumn{1}{c|}{71.5} & 
\multicolumn{1}{c|}{0.0} &
\multicolumn{1}{c||}{28.5} &
\multicolumn{1}{c|}{46.75} & 
\multicolumn{1}{c|}{0.0} &
\multicolumn{1}{c||}{53.25} &
\multicolumn{1}{c|}{5.5} &
\multicolumn{1}{c|}{0.0} &
\multicolumn{1}{c|}{94.5} \\ \hline

\multicolumn{1}{|c||}{K = 4} &
\multicolumn{1}{c|}{72.0} & 
\multicolumn{1}{c|}{0.0} &
\multicolumn{1}{c||}{28.0} &
\multicolumn{1}{c|}{47.0} & 
\multicolumn{1}{c|}{0.0} &
\multicolumn{1}{c||}{53.0} &
\multicolumn{1}{c|}{0.0} &
\multicolumn{1}{c|}{0.0} &
\multicolumn{1}{c|}{100.0} \\ \hline

\end{tabular}
\label{tab:k_analysis}
\end{table*}

\begin{table*}[t]
  \caption{Analysis of the proposed method's performance when interacting with reactive (IDM \cite{Treiber2000}) and predictive vehicles (CV, CV-Path and MPCC). }
         \centering
\begin{tabular}{c|c|c|c|c|c|c|c|c|c}
\hline
\multicolumn{1}{|c||}{}                 & \multicolumn{3}{|c||}{Cooperative} & \multicolumn{3}{c||}{Mixed} &
\multicolumn{3}{c|}{Non Cooperative} \\  \hline

\multicolumn{1}{|c||}{} &
\multicolumn{1}{c|}{Success(\%)} & \multicolumn{1}{c|}{Collision(\%)} &
\multicolumn{1}{c||}{Timeout(\%)} &
\multicolumn{1}{c|}{Success(\%)} &
\multicolumn{1}{c|}{Collision(\%)} &
\multicolumn{1}{c||}{Timeout(\%)} &
\multicolumn{1}{c|}{Success(\%)} &
\multicolumn{1}{c|}{Collision(\%)} &
\multicolumn{1}{c|}{Timeout(\%)} \\ \hline 

\multicolumn{1}{c}{React. Model} &
\multicolumn{1}{c}{} & \multicolumn{1}{c}{} &
\multicolumn{1}{c}{} &
\multicolumn{1}{c}{} &
\multicolumn{1}{c}{} &
\multicolumn{1}{c|}{} &
\multicolumn{1}{c}{} &
\multicolumn{1}{c}{} &
\multicolumn{1}{c}{} \\ \hline \hline  

\multicolumn{1}{|c||}{IDM \cite{Treiber2000}} &
\multicolumn{1}{c|}{86.0} & 
\multicolumn{1}{c|}{0.0} &
\multicolumn{1}{c||}{14.0} &
\multicolumn{1}{c|}{70.0} &
\multicolumn{1}{c|}{1.0} &
\multicolumn{1}{c||}{29.0} &
\multicolumn{1}{c|}{36.0} &
\multicolumn{1}{c|}{0.0} &
\multicolumn{1}{c|}{64.0} \\ \hline \hline 

\multicolumn{1}{c}{Pred. Model} &
\multicolumn{1}{c}{} & \multicolumn{1}{c}{} &
\multicolumn{1}{c|}{} &
\multicolumn{1}{c}{} &
\multicolumn{1}{c}{} &
\multicolumn{1}{c|}{} &
\multicolumn{1}{c}{} &
\multicolumn{1}{c}{} &
\multicolumn{1}{c}{} \\ \hline 

\multicolumn{1}{|c||}{CV} &
\multicolumn{1}{c|}{88.0} & 
\multicolumn{1}{c|}{0.0} &
\multicolumn{1}{c||}{12.0} &
\multicolumn{1}{c|}{72.0} & 
\multicolumn{1}{c|}{0.0} &
\multicolumn{1}{c||}{28.0} &
\multicolumn{1}{c|}{37.0} &
\multicolumn{1}{c|}{0.0} &
\multicolumn{1}{c|}{63.0}\\ \hline

\multicolumn{1}{|c||}{CV-Path} &
\multicolumn{1}{c|}{98.0} & 
\multicolumn{1}{c|}{0.0} &
\multicolumn{1}{c||}{2.0} &
\multicolumn{1}{c|}{88.0} & 
\multicolumn{1}{c|}{0.0} &
\multicolumn{1}{c||}{12.0} &
\multicolumn{1}{c|}{37.0} &
\multicolumn{1}{c|}{0.0} &
\multicolumn{1}{c|}{63.0} \\ \hline

\multicolumn{1}{|c||}{MPCC} &
\multicolumn{1}{c|}{89.0} & 
\multicolumn{1}{c|}{0.0} &
\multicolumn{1}{c||}{11.0} &
\multicolumn{1}{c|}{76.0} & 
\multicolumn{1}{c|}{0.0} &
\multicolumn{1}{c||}{24.0} &
\multicolumn{1}{c|}{39.0} &
\multicolumn{1}{c|}{0.0} &
\multicolumn{1}{c|}{61.0} \\ \hline

\end{tabular}
\label{tab:diff_pred_models}
\end{table*}

The MPCC's parameters (i.e., weights and velocity reference) highly influence the local planner's performance. 
Here, we study the two key components controlling the AV's interaction with the other vehicles: the velocity tracking weight ($q_v$) and the reference velocity ($v_{\textrm{ref}}$). \cref{tab:mpcc_weigth} presents performance results for different $q_v$ and $v_{\textrm{ref}}$ values. Increasing the reference velocity combined with high $q_v$ values generates more aggressive behavior and significantly reduces the timeout rate. However, it also increases the collision rate. In contrast, low $q_v$ values weaken the influence of the velocity reference on the MPCC performance. The presented results demonstrate that fine-tuning the MPCC's weights and velocity reference is insufficient for safe and efficient navigation in dense traffic environments, supporting the need for an interaction-aware velocity reference. $q_v=1.0$ and $v_{\textrm{ref}}= 2$ m/s lead to the best performance, i.e., higher success rate and lower collision and timeout rate. For the following experiments, we use $q_v = 1.0$ and a velocity reference of $v_{\textrm{ref}} =$ 2 m/s for the MPCC baseline.

\subsubsection{Hyperparameter selection}\label{sec:ablation_k}

A key design choice of the proposed framework is the number of control cycles per policy query, denoted by $K$. For instance, for $K=1$, we query the policy network for a new velocity reference for each control cycle, while for $K=4$,  we use the same queried velocity reference during $4$ control cycles. Here, we study the impact on the learned policy's performance for $K=\{1,\dots,4\}$. During testing, all the policies are evaluated using $K = 1$. \cref{tab:k_analysis} summarizes the obtained performance results. The policy trained with $K = 2$ outperforms the other policies in terms of success and collision rate. The policy trained with $K = 1$ elicits an overly aggressive response from AV, evident from a high collision rate and a low timeout percentage. In contrast, higher $K$ values lead the AV to exhibit an overly conservative behavior, thus, higher timeout percentage. This behavior can be attributed to the long duration for which the same action is applied after querying the interactive policy. For instance, using a large velocity reference value during many control cycles highly increases the collision likelihood at the merging point. This compels the RL algorithm to learn biased policy towards low-velocity references to avoid an impending collision resulting in an overly conservative behavior. Finally, the policy trained with $K = 2$ elicits a balanced response from the AV that is neither too conservative nor too aggressive, resulting in a high success rate and a low collision rate for all the scenarios.

\subsubsection{Simulation environment}\label{sec:abl_simulation_models} 

This work introduces an IDM variant enhancing the other vehicles with anticipatory behavior. Our proposed model (P-IDM in \cref{sec:other_agents_model}) relies on the assumption that the other vehicles can infer the AV's motion plans. Here, we evaluate the influence of the prediction model used to infer the AV's plans on our method's performance. We consider the following prediction models variants:

\begin{enumerate}
    \item CV: Constant velocity (CV) model;
    \item CVPath: Constant velocity (CV) model along the AV's reference path;
    \item MPCC: MPCC plan (\cref{eq:mpc_formulation}) assuming the AV's current velocity as the velocity reference, $v_{\textrm{ref}}=v_k$.
\end{enumerate}
Moreover, we also evaluate our method's performance in reactive scenarios employing the IDM \cite{Treiber2000} to model the other vehicles' behaviors. The interactive policy was trained considering a mixed setting of other vehicles following a P-IDM model with CV predictions. The presented results in \cref{tab:diff_pred_models} demonstrate that our proposed approach is robust and generalizes well to environments with other vehicles exhibiting different behaviors. Employing the CV-Path prediction model results in highly cooperative behavior for other vehicles as shown by the high success rate. In contrast, the scenarios with vehicles following an IDM \cite{Treiber2000} represents the most challenging scenario.

\subsection{Qualitative Results} \label{sec:qualitative}

\begin{figure*}
  \centering
  \begin{subfigure}{\textwidth}
  
    \includegraphics[height=1.6cm,width=18cm,trim={0cm 0cm 0cm 0cm},clip]{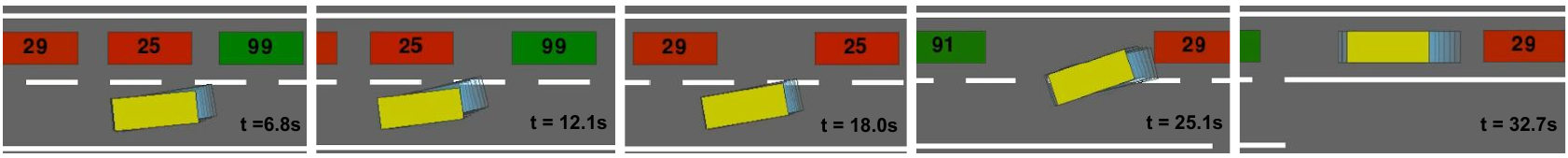}
        \caption{ Successful merging maneuver: As the AV approaches the merging point, it tries to assess the reaction of its action on the vehicle titled "25" by inching closer to the main lane. The vehicle's non-cooperative behavior does not elicit a response typical of vehicles willing to yield, forcing the AV to stop. It tries the same with the vehicle titled "29" by creeping closer to the main lane but fails again. Finally, the merge is successful when a cooperative vehicle titled "91" emerges and gives way to the AV. }
    \label{fig:merging-scenario}
  \end{subfigure}

    \hspace{10mm}
    \begin{subfigure}{\textwidth}
    
      \includegraphics[height=1.6cm,width=18cm,trim={0cm 0cm 0cm 0cm},clip]{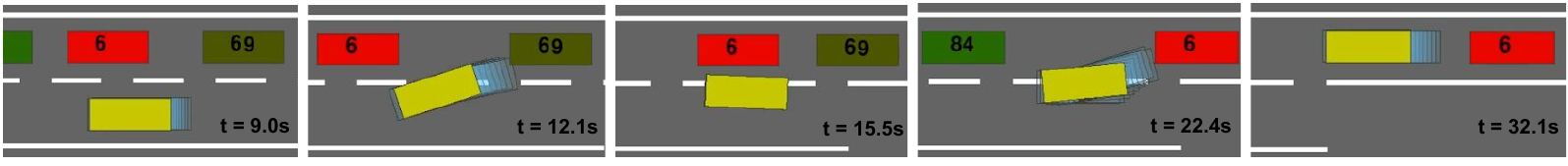} 
      \caption{Attempting to merge with a non-cooperative vehicle: In this episode, the guidance policy wrongfully estimates the other vehicle's non-cooperative nature, titled "6", compelling the AV to merge in front of the other agent. However, the obstacle avoidance constraint forces the AV to steer away from the other vehicle to avoid a collision. Finally, the AV merges in front of the cooperative vehicle titled "84".  }
      \label{fig:collision-avoidance-scenario}
    \end{subfigure}
    
    \vspace{1mm}
    \begin{subfigure}{\textwidth}
      \includegraphics[height=1.8cm,width=18cm,trim={0cm 0cm 0cm 0cm},clip]{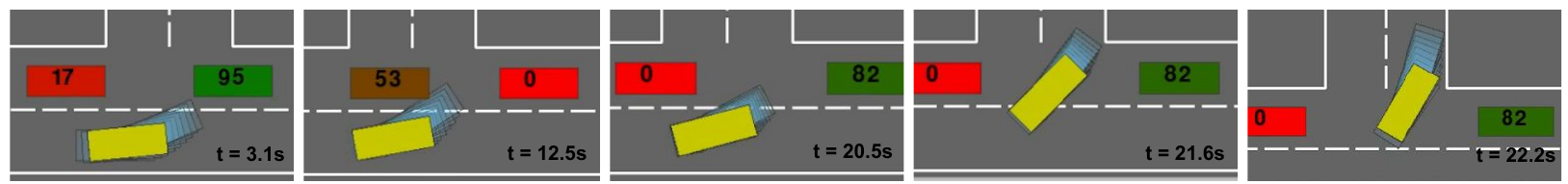}
      \caption{
      Unprotected left-turn scenario: the AV approaches the center of the main lane to make the other vehicles yield. The first three vehicles it meets are non-cooperative and do not stop. When it meets a cooperative vehicle, titled "82", the AV behavior induces the other vehicle to yield allowing the AV to cross successfully. 
      }
     \label{fig:left-turn-scenario}
    \end{subfigure}
    \begin{subfigure}{\textwidth} 
    
    \centering
      \includegraphics[height=0.6cm,width=14cm,trim={0cm 0cm 0cm 0cm},clip]{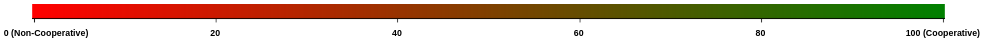}
    \end{subfigure}

  \caption{ All the scenarios employ the P-IDM model (\cref{sec:other_agents_model}) to simulate the other vehicles. The AV is represented in yellow, whereas the future states, as computed by the MPCC, are plotted in light blue. Each other vehicle is assigned with a color transitioning from red (i.e., non-cooperative) to green (i.e., cooperative) to highlight the other vehicles' cooperativeness. The number displayed by each other vehicle represents its cooperation coefficient. All the numbers in between show a continuous transition from non-cooperative (0) to cooperative (100).
  }
  \label{fig:learned_behavior}
\end{figure*}

\cref{fig:learned_behavior} presents visual results for our method for the merging and left-turn scenarios. 
In Fig. \ref{fig:merging-scenario}, the AV successfully merged onto the main lane by leveraging other vehicles' cooperativeness. In contrast, in Fig. \ref{fig:collision-avoidance-scenario}, we highlight a critical advantage of our framework: the ability to perform a collision avoidance maneuver when the guidance policy wrongly estimates the other vehicle's cooperativeness. In this episode, at 12.1 s, the AV initiates a merging maneuver. However, the non-cooperative vehicle does not allow it. The local planner aborts and starts a collision avoidance maneuver at 15.5 s, merging successfully later when encountering a cooperative vehicle at 22.4 s.
Finally, Fig. \ref{fig:merging-scenario} shows the AV performing an unprotected left-turn maneuver successfully. The presented qualitative results show that our proposed method enables the AV to safely and efficiently navigate in dense traffic scenarios. We refer the reader to the video accompanying this paper for more qualitative results.

\begin{table*}
  \caption{ Statistic results for 1200 runs of proposed method (IntMPC) compared to baselines (MPCC~\cite{ferranti2019safevru} and DRL~\cite{Haarnoja2018}) considering three different settings for the other vehicles (\cref{sec:scenarios}): percentage of success, collisions and timeout episodes.}
         \centering
\begin{tabular}{c|c|c|c|c|c|c|c|c|c}
\hline
\multicolumn{1}{|c||}{}                 & \multicolumn{3}{|c||}{Cooperative} & \multicolumn{3}{c||}{Mixed} &
\multicolumn{3}{c|}{Non Cooperative} \\  \hline

\multicolumn{1}{|c||}{} &
\multicolumn{1}{c|}{Success(\%)} & \multicolumn{1}{c|}{Collision(\%)} &
\multicolumn{1}{c||}{Timeout(\%)} &
\multicolumn{1}{c|}{Success(\%)} &
\multicolumn{1}{c|}{Collision(\%)} &
\multicolumn{1}{c||}{Timeout(\%)} &
\multicolumn{1}{c|}{Success(\%)} &
\multicolumn{1}{c|}{Collision(\%)} &
\multicolumn{1}{c|}{Timeout(\%)} \\ \hline \hline 
\multicolumn{1}{|c||}{MPCC \cite{ferranti2019safevru}} &
\multicolumn{1}{c|}{78.0} & 
\multicolumn{1}{c|}{11.0} &
\multicolumn{1}{c||}{10.0} &
\multicolumn{1}{c|}{62.0} & 
\multicolumn{1}{c|}{22.0} &
\multicolumn{1}{c||}{16.0} &
\multicolumn{1}{c|}{26.0} &
\multicolumn{1}{c|}{57.0} &
\multicolumn{1}{c|}{19.0} \\ \hline


\multicolumn{1}{|c||}{RL \cite{Haarnoja2018}} &
\multicolumn{1}{c|}{86.0} & 
\multicolumn{1}{c|}{3.0} &
\multicolumn{1}{c||}{11.0} &
\multicolumn{1}{c|}{69.0} &
\multicolumn{1}{c|}{2.0} &
\multicolumn{1}{c||}{29.0} &
\multicolumn{1}{c|}{31.0} &
\multicolumn{1}{c|}{5.0} &
\multicolumn{1}{c|}{64.0} \\ \hline


\multicolumn{1}{|c||}{Int-MPC} &
\multicolumn{1}{c|}{\textbf{86.0}} & 
\multicolumn{1}{c|}{\textbf{0.0}} &
\multicolumn{1}{c||}{14.0} &
\multicolumn{1}{c|}{\textbf{70.0}} & 
\multicolumn{1}{c|}{\textbf{0.0}} &
\multicolumn{1}{c||}{30.0} &
\multicolumn{1}{c|}{\textbf{36.0}} &
\multicolumn{1}{c|}{\textbf{0.0}} &
\multicolumn{1}{c|}{64.0} \\ \hline

\end{tabular}
\label{tab:main results}
\end{table*}

\subsection{Quantitative Results} \label{sec:quantitative}

Aggregated results in \cref{tab:main results} show that our method outperforms the baseline methods in terms of successful merges and number of collisions considering different settings for the other vehicles' behaviors (i.e., cooperative, mixed and, non-cooperative). The combined capability of interactive RL policy to implicitly embed inter-vehicle interactions into the velocity's policy and the safety provided by the collision avoidance constraints allows our method to succeed in all the environments. The optimization-based baseline (MPCC) shows poor performance for all settings, i.e., high collision rate. The reason is the lack of assimilation of inter-vehicle interactions into the policy and a tracking velocity reference error term in the cost function formulation that motivates the AV to keep the same velocity disregarding the nearby vehicles' cooperativeness. 
The DRL baseline achieves significantly higher performance, i.e., lower collision rate and a higher number of successful episodes. Nevertheless, it still leads to a significant number of collisions due to the lack of collision avoidance constraints to ensure safety when closely interacting with other vehicles. This demonstrates that employing collision constraints for navigation in dense traffic scenarios leads to superior performance over solely learning-based methods.

\cref{tab:time_to_goal} presents statistical results of the \textit{time-to-goal} for all methods. To evaluate the statistical significance, we performed pairwise Mann–Whitney U-tests between each method, considering a 95$\%$ confidence level. The results show statistical significance for the MPCC's results against the other methods for cooperative and mixed settings. In contrast, there is no statistical difference in terms of \textit{time-to-goal} between the DRL and IntMPC. Similarly, between all methods in non-cooperative environments. The presented results show that employing collision avoidance constraints do not increase the average \textit{time-to-goal} while improving safety. Moreover, in non-cooperative environments, all methods achieve comparable performance in terms of \textit{time-to-goal}.

\begin{table}[t]

\begin{center}
\caption{ Statistical results on the \textit{time-to-goal} [s]. Only the episodes where all methods are successful are considered in the presented results. Bold values represent the results with statistical significance.}
\label{tab:time_to_goal}
\begin{tabular}{l c c c}

\hline
 & Cooperative & Mixed & Non-cooperative  \\
 \hline
 MPCC \cite{ferranti2019safevru}  & \textbf{34.7} $\pm$ \textbf{4.1}  &  \textbf{35.9} $\pm$ \textbf{6.9}  &  40.1 $\pm$ 5.8  \\
 DRL \cite{Haarnoja2018} &  37.5 $\pm$ 7.8  &  37.9 $\pm$ 6.9  &  41.8 $\pm$ 7.4  \\
 IntMPC & 37.6 $\pm$ 8.0 & 37.7 $\pm$ 6.0 & 41.0 $\pm$ 7.3 \\
 \hline
 \end{tabular}
 \end{center}
\end{table}

\begin{figure}[t!]
\centering
\includegraphics[scale=0.32]{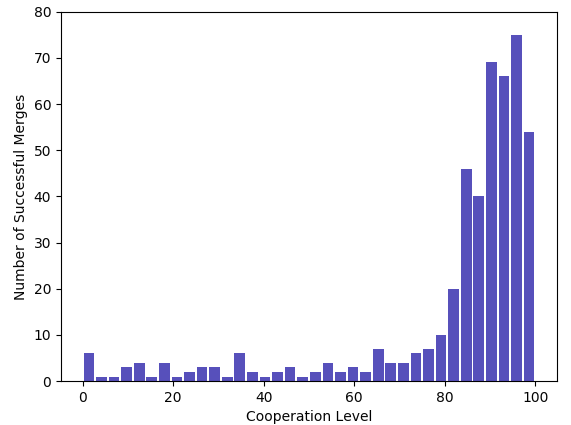}

\caption{This figure provides a comprehensive analysis of the agents' cooperation level (0 - non cooperative, 100 - cooperative) in front of which the ego vehicle was able to merge successfully. }
\label{fig:hist}
\end{figure}

To demonstrate our policy's ability to leverage agents' cooperativeness explicitly, we evaluate 600 episodes in a mixed scenario where we track the other vehicle' cooperation level in front of which the AV performs a successful merging maneuver. Fig. \ref{fig:hist} depicts a histogram illustrating the number of successful episodes per cooperation coefficient, demonstrating that our method mostly merges with cooperative vehicles. A small number of successful merges can be seen with non-cooperative vehicles as well. This behavior can be attributed to the random sampling of IDM parameters resulting in different agents' acceleration values. Thus, the agents might leave a gap big enough for the AV to merge onto the lane when moving from a standstill position.

\begin{figure}[!t]
\centering
\includegraphics[scale=0.15]{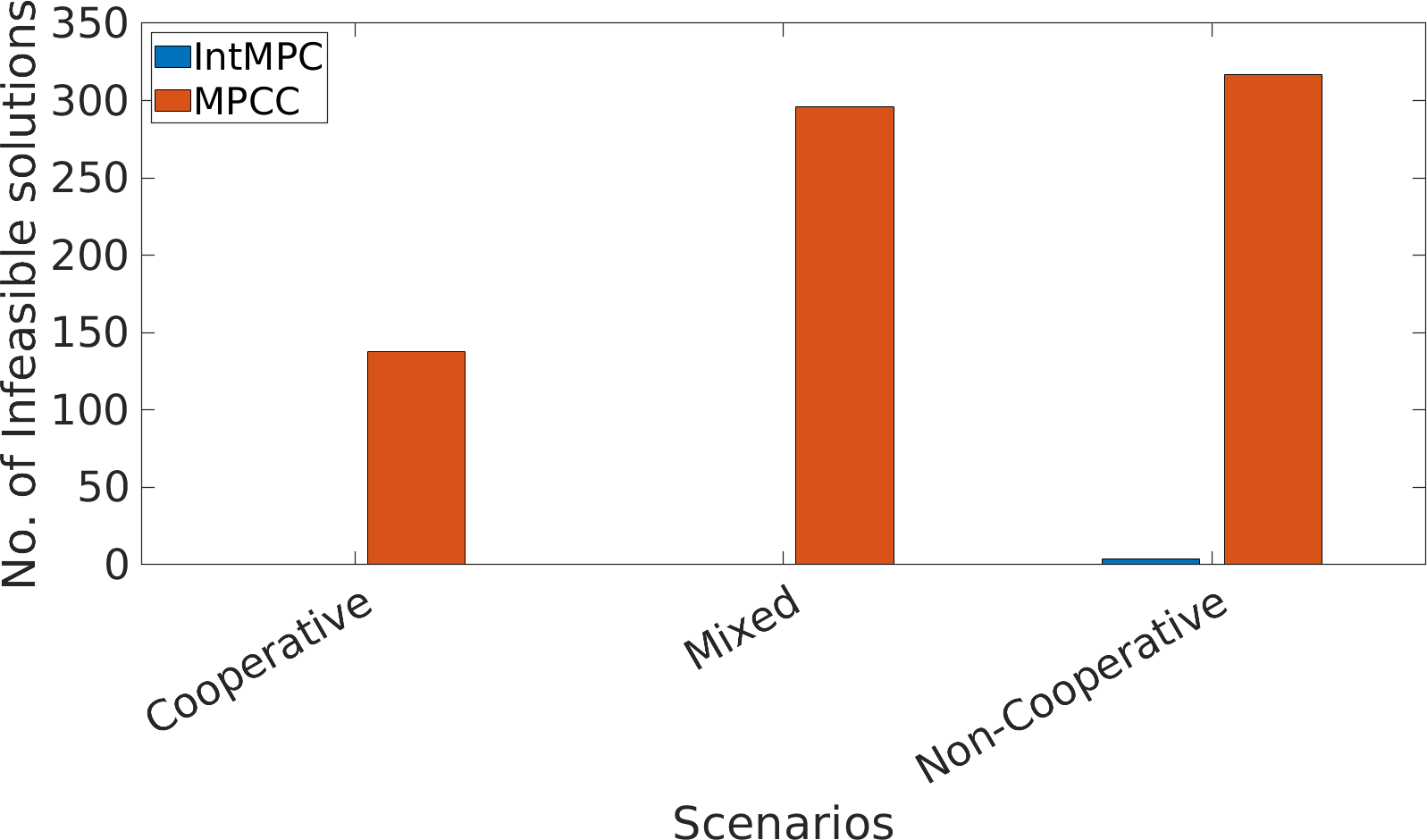}

\caption{Number of infeasible solutions encountered by the solver for our method (IntMPC) versus the optimization-based baseline (MPCC).}
\label{fig:policy_bar}
\end{figure}

\cref{fig:policy_bar} presents the number of infeasible solutions for our method (IntMPC) and the MPCC baseline. To jointly train the RL policy with the local controller and penalize the state and action tuples resulting in the solver infeasibility, significantly reduces the number of infeasible solutions.
Finally, in terms of computation performance, our policy's network has an average computation time of $1.35 \pm 0.5$ ms. To solve the IntMPC's optimization problem (\cref{eq:mpc_formulation}) takes on average $3.0 \pm 1.35$ ms for all experiments. There was no statistical difference on the policy's and solver's computation times for the different settings of the other vehicles (e.g., cooperative, mixed and non-cooperative). These results demonstrate out method's real-time applicability.

\section{Conclusion}
This paper introduced an interaction-aware policy for guiding a local optimization planner through dense traffic scenarios. We proposed to model the interaction policy as a velocity reference and employed DRL methods to learn a policy maximizing long-term rewards by exploiting the interaction effects. Then, a MPCC is used to generate control commands satisfying collision and kino-dynamic constraints when a feasible solution is found. Learning an interaction-aware velocity reference policy enhances the MPCC planner with interactive behavior necessary to safely and efficiently navigate in dense traffic.
The presented results show that our method outperforms solely learning-based and optimization-based planners in terms of collisions, successful maneuvers, and fewer deadlocks in cooperative, mixed, and non-cooperative scenarios. 

As future works, we plan to replace the simple constant velocity model with an interaction-aware prediction model learned from data \cite{brito2020social}. This will improve the prediction performance significantly and so, safety and performance. We intend to expand our framework to provide local guidance on the heading direction for the AV and evaluate it in more unconstrained scenarios, such as lane-changing in highways. Finally, we plan to implement and evaluate our method in a real autonomous vehicle.





\ifCLASSOPTIONcaptionsoff
  \newpage
\fi

\bibliographystyle{IEEEtran}
\bibliography{IEEEabrv,main.bib}

\begin{IEEEbiography}[{\includegraphics[width=1in,height=1.25in,clip,keepaspectratio]{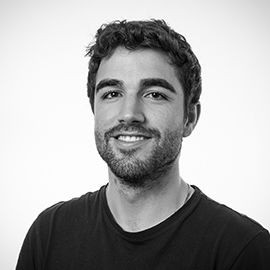}}]{Bruno Brito}
Bruno Brito is a Ph.D. student at the Department of Cognitive Robotics at the Delft University of Technology. He received the M.Sc. (2013) degree from the Faculty of Engineering of the University of Porto. Between 2014 and 2016, he was a trainee at the European Space Agency (ESA) in the Guidance, Navigation and Control section. After, he was a Research Associate, between 2016 and 2018, in the Fraunhofer Institute for Manufacturing Engineering and Automation. Currently, his research is focused on developing motion planning algorithms bridging learning-based and optimization-based methods for autonomous navigation among humans.
\end{IEEEbiography}

\begin{IEEEbiography}[{\includegraphics[width=1in,height=1.25in,clip,keepaspectratio]{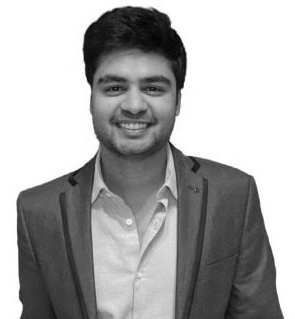}}]{Achin Argawal}
Achin Agarwal received the M.Sc. degree in Mechanical Engineering from the Delft University of Technology in December 2020,  with a Master Thesis focused on modeling highly interactive behavior between various traffic entities and navigation in dense traffic. Currently, he is exploring the application of Artificial Intelligence techniques for modeling the influence of various parameters involved in financial markets. 

\end{IEEEbiography}

\begin{IEEEbiography}[{\includegraphics[width=1in,height=1.25in,clip,keepaspectratio]{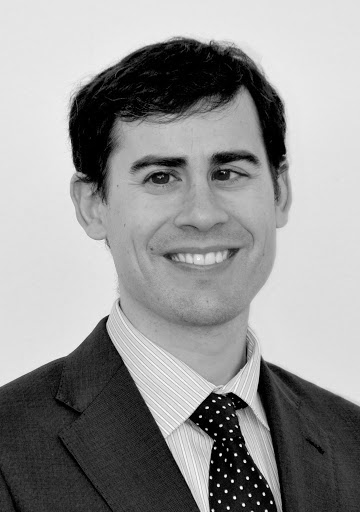}}]{Javier Alonso-Mora}

Javier Alonso-Mora is an Associate Professor at the Delft University of Technology and head of the Autonomous Multi-robots Laboratory. Dr. Alonso-Mora was a Postdoctoral Associate at MIT and received his Ph.D. degree from ETH Zurich, where he worked in a partnership with Disney Research Zurich. His main research interest is in navigation, motion planning and control of autonomous mobile robots, with a special emphasis on multi-robot systems, on-demand transportation and robots that interact with other robots and humans in dynamic and uncertain environments. He is the recipient of multiple prizes and grants, including the ICRA Best Paper Award on Multi-robot Systems (2019), an Amazon Research Award (2019) and a talent scheme VENI award from the Netherlands Organisation for Scientific Research (2017). 
\end{IEEEbiography}


\end{document}